\documentclass[conference]{IEEEtran}
\IEEEoverridecommandlockouts
\usepackage{cite}
\usepackage{amsmath,amssymb,amsfonts}
\usepackage{algorithmic}
\usepackage{graphicx}
\usepackage{textcomp}
\usepackage{xcolor}
\usepackage{times}
\usepackage{epsfig}
\usepackage{amsmath}
\usepackage{amssymb}
\usepackage{multirow}
\usepackage{booktabs}

\def\BibTeX{{\rm B\kern-.05em{\sc i\kern-.025em b}\kern-.08em
    T\kern-.1667em\lower.7ex\hbox{E}\kern-.125emX}}
\begin{document}

\title{PhysLayer: Language-Guided Layered Animation with Depth-Aware Physics}

\author{
\IEEEauthorblockN{Tianyidan Xie\textsuperscript{1},
Zhentao Huang\textsuperscript{2},
Mingjie Wang\textsuperscript{3},
Xin Huang\textsuperscript{3},
Jun Zhou\textsuperscript{4},
Minglun Gong\textsuperscript{2},
Zili Yi\textsuperscript{1,*}\thanks{*Corresponding author: yi@nju.edu.cn}}
\IEEEauthorblockA{\textsuperscript{1}Nanjing University \quad
\textsuperscript{2}University of Guelph \quad
\textsuperscript{3}Zhejiang Sci-Tech University \quad
\textsuperscript{4}Dalian Maritime University}
}

\maketitle

\begin{abstract}
Existing image-to-video generation methods often produce physically implausible motions and lack precise control over object dynamics. While prior approaches have incorporated physics simulators, they remain confined to 2D planar motions and fail to capture depth-aware spatial interactions. We introduce PhysLayer, a novel framework enabling language-guided, depth-aware layered animation of static images. PhysLayer consists of three key components: First, a language-guided scene understanding module that utilizes vision foundation models to decompose scenes into depth-based layers by analyzing object composition, material properties, and physical parameters. Second, a depth-aware layered physics simulation that extends 2D rigid-body dynamics with depth motion and perspective-consistent scaling, enabling more realistic object interactions without requiring full 3D reconstruction. Third, a physics-guided video synthesis module that integrates simulated trajectories with scene-aware relighting for temporally coherent results. Experimental results demonstrate improvements in CLIP-Similarity (+2.2\%), FID score (+9.3\%), and Motion-FID (+3\%), with human evaluation showing enhanced physical plausibility (+24\%) and text-video alignment (+35\%). Our approach provides a practical balance between physical realism and computational efficiency for controllable image animation.
\end{abstract}


\section{Introduction}

The advent of advanced diffusion models has revolutionized image-to-video generation, leading to unprecedented improvements in quality and performance~\cite{blattmann2023stable, chen2023seine}. Despite significant advancements, a fundamental challenge remains: generating videos that not only maintain high visual fidelity but also exhibit physical realism, accurately adhering to real-world physics while allowing precise control over object motion and interactions. This challenge is particularly crucial for applications in content creation, visual effects, and interactive media, where both visual quality and physical accuracy are indispensable. Current methods generally fall into two categories: those that rely solely on generative models, which often produce unnatural motions that violate physical laws \cite{xing2025dynamicrafter, zhang2023i2vgen, yang2024cogvideox}, and those that integrate physics simulations but are constrained to 2D planar dynamics with limited controllability \cite{liu2025physgen}.

Specifically, we identify three key challenges in existing methods: First, they often require complex motion parameters or generate unnatural motions that violate physical laws. Second, while some approaches incorporate physics simulation, they are typically confined to 2D planar dynamics, failing to capture the rich spatial interactions inherent in real-world 3D scenes. Third, current methods lack a unified framework that enables flexible and precise control over diverse animation tasks using both language guidance and visual control.

\begin{figure}[t]
    \begin{center}
    \includegraphics[width=1.0\linewidth]{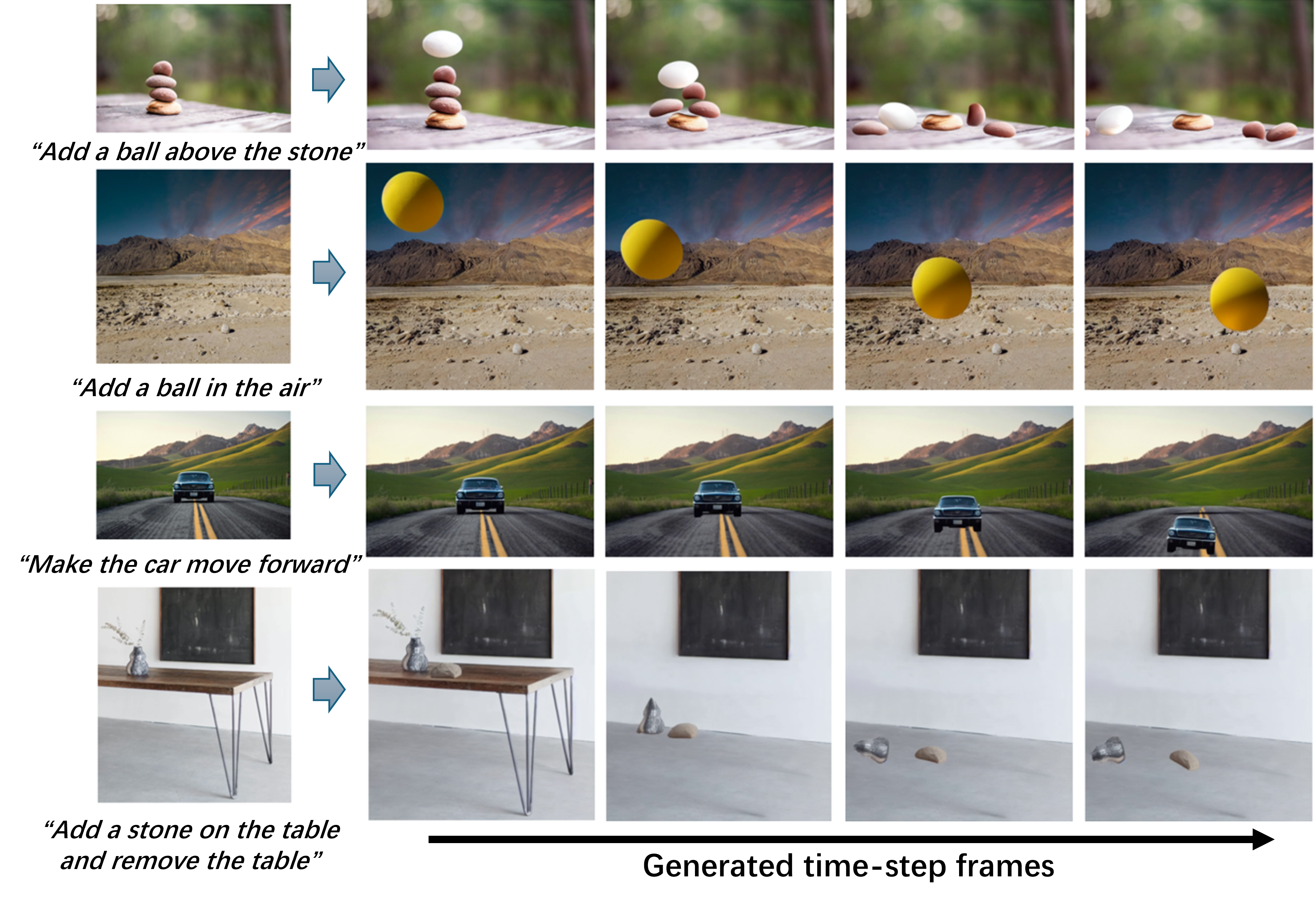}
    \end{center}
    \caption{Visualizing PhysLayer's capabilities across diverse animation scenarios. Our method supports (a) object insertion with depth-aware collision handling (rows 1 and 2); (b) controlled object manipulation with accurate perspective scaling (row 3); and (c) consistent, physically aware scene reconstruction (row 4). Across all examples, our approach maintains both physical plausibility and visual coherence throughout the generated sequence.}
    \label{fig:teaser}
    \end{figure}

To overcome these challenges, we propose PhysLayer, a novel framework that extends physics-simulation methods into the 3D domain and enables language-guided animation of static images. Our approach is structured into a three-phase procedure:

First, Language-Guided Scene Understanding and Layer Decomposition: We leverage large vision foundation models and advanced image parsing techniques for comprehensive scene understanding. This involves reasoning about object composition, material properties, and physical parameters. Based on depth information, we decompose the scene into multiple interactive layers, capturing spatial relationships and enabling individual object manipulation.

Second, Depth-Aware Physics Simulation: We extend traditional 2D rigid-body physics into 3D space through depth-aware multi-planar dynamics simulation. This allows us to simulate object dynamics with physical accuracy, accounting for interactions between objects and their environment across varying depth planes. By incorporating depth variations, we enable realistic simulations of object-object and object-scene interactions that reflect real-world physics.

Third, Physics-Guided Video Synthesis Pipeline: We design a video synthesis pipeline that seamlessly integrates the physically simulated trajectories with scene-aware relighting. By utilizing diffusion-based refinement and rendering techniques, we enhance visual quality and temporal coherence. This results in visually compelling videos that maintain both physical plausibility and high aesthetic appeal.

Our contributions can be summarized as:

\begin{itemize}
    \item \textbf{A Language-Guided Depth-Aware Animation Framework:} We present a framework that offers flexible and precise control over static image animation through natural language instructions. By decomposing scenes into depth-based layers and reasoning about physical properties, our approach enables intuitive animation control while maintaining physical plausibility.
    
    \item \textbf{A Layered Depth-Aware Physics Simulation:} We introduce a depth-aware physics simulation technique that extends traditional 2D rigid-body dynamics with depth motion and perspective-consistent scaling. By operating on discretized depth layers, our method captures more realistic spatial interactions than pure 2D methods while avoiding the complexity of full 3D reconstruction. This provides a practical trade-off between physical realism and computational efficiency.
    
    \item \textbf{Comprehensive Evaluation:} Extensive experiments demonstrate PhysLayer's effectiveness across multiple metrics. We achieve improvements in CLIP-Similarity (2.2\%), FID score (9.3\%), and Motion-FID (3\%). Human evaluation validates substantial gains in physical plausibility (24\%) and text-video alignment (35\%). We provide ablation studies and physical consistency analysis to validate each component's contribution.
\end{itemize}

\section{Related Works}

\subsection{Large Model Based Physical Reasoning}

Visual physical reasoning—the ability to understand and predict physical phenomena from visual observations—has long been a challenging problem in computer vision and AI. Traditional approaches often depend on specialized physics simulators or hand-crafted rules~\cite{wu2016physics, wu2017learning}. Early benchmarks like CLEVRER~\cite{yi2019clevrer} and PHYRE~\cite{bakhtin2019phyre} focused on reasoning about collision events and physical interactions.

The advent of large foundation models has revolutionized zero-shot capabilities in physical reasoning. GPT-4V~\cite{achiam2023gpt} exhibits impressive abilities in inferring physical properties and predicting object interactions without explicit physics training. Recent studies have leveraged these capabilities for various physical tasks: PAC-NeRF~\cite{li2023pac} utilizes Vision-Language Models (VLMs) for physics-aware scene reconstruction, while~\cite{zhai2024physical} employs language-embedded feature fields for understanding physical properties.

Beyond direct inference, large models show promise in guiding physics simulations. The Dynamic Concept Learner~\cite{chen2021grounding} integrates visual, linguistic, and physics-based reasoning to enhance dynamic predictions. Recent work, PhysGen~\cite{liu2025physgen}, illustrates the potential of using vision-language models for physical property estimation through multiple object-centric queries. Our method builds on this concept by performing holistic scene understanding in a single pass.

\subsection{Image-based Physical Simulation}

Physical simulation from visual inputs has been extensively studied in computer vision and graphics. Traditional approaches rely on specialized physics engines~\cite{blomqvistpymunk, coumans2021pybullet} for rigid body dynamics, fluid simulation, and deformable objects. While full 3D simulation provides accurate physics, it requires complete 3D scene reconstruction and understanding, which remains challenging from a single image.

Recent work, PhysGen~\cite{liu2025physgen}, demonstrates using vision-language models for physical property estimation through object-centric queries. However, PhysGen operates purely in 2D space without depth awareness, limiting its ability to model perspective changes and depth-based interactions. Our method builds on this concept by performing holistic scene understanding in a single pass while extending the simulation to handle depth-aware dynamics, providing a more realistic representation of spatial relationships.

\subsection{Video Generation and Image Animation}

Video generation has witnessed remarkable progress with the emergence of diffusion models~\cite{ho2020denoising}. Recent large-scale text-to-video models like I2VGen-XL~\cite{zhang2023i2vgen} and SEINE~\cite{chen2023seine} demonstrate impressive capabilities in open-domain video synthesis through cascaded diffusion and temporal modeling. CogVideoX~\cite{yang2024cogvideox} and DynamiCrafter~\cite{xing2025dynamicrafter} further improve generation quality by introducing expert transformers and specialized video diffusion priors.

Image animation offers a more controlled approach to video synthesis. Traditional methods focus on bringing static images to life through motion synthesis~\cite{holynski2021animating}. Recent works like PIA~\cite{zhang2024pia} introduce plug-and-play modules to achieve personalized image animation. However, most existing methods operate primarily in 2D image space, lacking the ability to model depth-aware motion and spatial interactions accurately.

Our method addresses these limitations by combining physics-based simulation with modern diffusion models. By explicitly modeling depth-aware object dynamics and interactions, we achieve more precise control over object movements in both planar and depth directions.

\begin{figure*}[t]
\begin{center}
\includegraphics[width=0.95\textwidth]{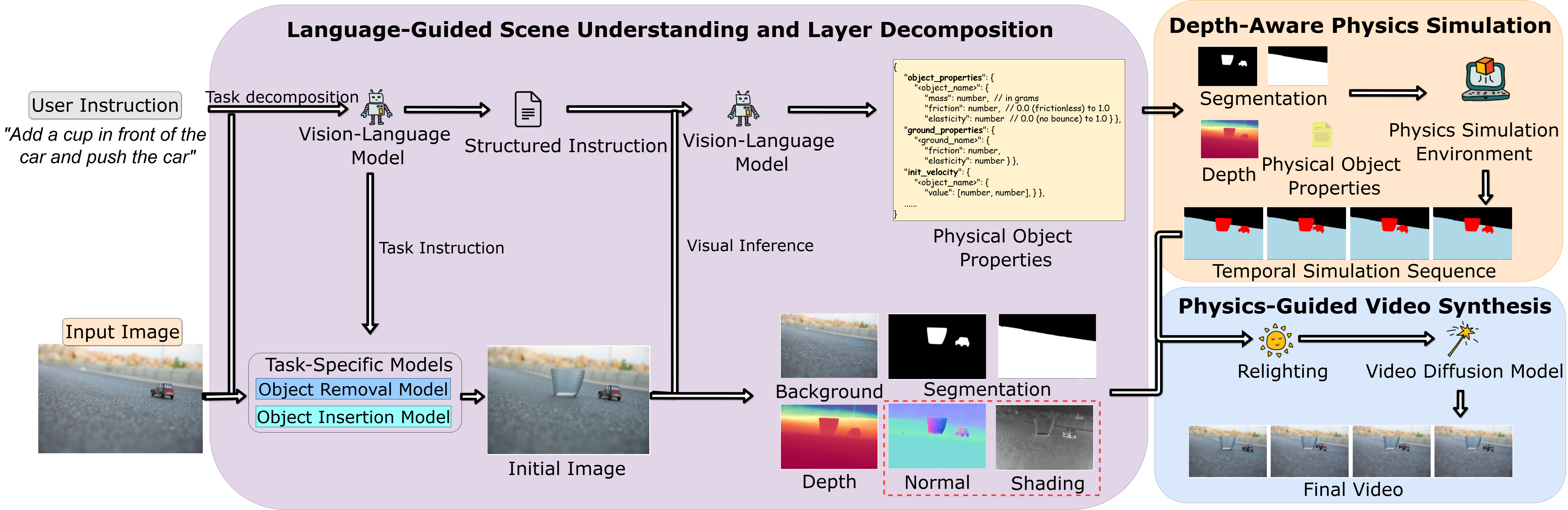}
\end{center}
\caption{\textbf{PhysLayer framework}. Our language-guided image animation framework consists of three components: (1) Language-Guided Scene Understanding and Layer Decomposition, (2) Depth-Aware Physics Simulation, and (3) Physics-Guided Video Synthesis.}
\label{fig:method}
\end{figure*}

\section{Method}

Given a single input image and natural language instructions, our goal is to generate a realistic video that features physically plausible dynamics while adhering to the instructed animation task. Our key insight is to bridge natural language understanding with depth-aware physics simulation through a three-stage framework.

\subsection{Language-guided Scene Understanding and Layer Decomposition}

The language-guided scene understanding and layer decomposition module generates a comprehensive scene representation by jointly reasoning about semantic, geometric, and physical properties through scene understanding and visual inference.

\noindent\textbf{Scene Understanding:} Our framework leverages large vision-language models (VLMs)~\cite{achiam2023gpt, team2023gemini} to develop a thorough understanding of user intentions and scene physics. This process is achieved through two primary analyses: (1) \textit{Task Decomposition}: The VLM interprets natural language instructions to identify animation objectives and associated physical interactions. Based on this analysis, animations are categorized into three core operations: object insertion, removal, and control. (2) \textit{Physical Property Inference}: Building on recent advances in language-driven physical reasoning~\cite{gao2024physically, zhai2024physical}, the VLM infers key physical properties such as mass, density, friction coefficients, and elasticity.

\noindent\textbf{Vision Inference:} Our framework integrates multiple specialized vision models to build a detailed and accurate scene representation. We employ Grounded-SAM~\cite{ren2024grounded} for instance-level semantic segmentation, which accurately delineates objects, boundaries, and supporting surfaces. Spatial configurations and depth relationships are captured using GeoWizard~\cite{fu2025geowizard}, a state-of-the-art depth and normal estimation network that enables depth-aware physical simulation.

To achieve photorealistic rendering, we utilize an intrinsic decomposition model~\cite{careaga2023intrinsic} to analyze scene illumination conditions. For static scene structure reconstruction, we apply advanced generative inpainting techniques~\cite{rombach2022high, ekin2024clipaway}, which recover complete background content in regions initially occupied by dynamic objects.

\subsection{Depth-aware Physics Simulation}

Building on the physical properties and depth information extracted during scene understanding, we introduce a depth-aware layered physics simulation framework. Unlike previous methods~\cite{liu2025physgen} that operate exclusively in 2D space, our approach extends traditional rigid-body dynamics with depth motion constraints and perspective-consistent scaling. Rather than requiring full 3D reconstruction, we achieve depth awareness through a layered representation that discretizes the depth dimension while maintaining computational efficiency.

\paragraph{Depth Layer Representation} We discretize the scene depth into $L$ layers based on the estimated depth map from GeoWizard~\cite{fu2025geowizard}. Each layer $l$ corresponds to a depth range $[d_l, d_{l+1}]$, where objects within the same range interact physically. The number of layers is determined by analyzing the depth histogram and clustering objects at similar depths: $L = \min(N_{obj} / 2, 8)$, where $N_{obj}$ is the number of dynamic objects. This layered representation enables efficient collision detection while approximating depth-aware interactions.

We formulate our depth-aware dynamics in an augmented 2.5D representation. At time $t$, each rigid object $i$ is represented by a state vector including translation $\mathbf{t}_i(t) \in \mathbb{R}^3$ (where the third dimension represents depth displacement) and rotation $\theta_i(t) \in SO(2)$ constrained to the image plane. The velocity state consists of linear velocity $\mathbf{v}_i(t) \in \mathbb{R}^3$ and angular velocity $\omega_i(t) \in \mathbb{R}$.

The complete state of object $i$ at time $t$ is formally expressed as $\mathbf{q}_i(t) = [\mathbf{t}_i(t), \theta_i(t), \mathbf{v}_i(t), \omega_i(t)]$, with dynamics governed by:

\begin{equation}
\frac{d}{dt}\mathbf{q}(t) = \frac{d}{dt}\begin{bmatrix}
\mathbf{t}(t) \\
\theta(t) \\
\mathbf{v}(t) \\
\omega(t)
\end{bmatrix} = \begin{bmatrix}
\mathbf{v}(t) \\
\omega(t) \\
\frac{\mathbf{F}_{3D}(t)}{M} \\
\frac{\tau(t)}{I}
\end{bmatrix}
\end{equation}

Here, $\mathbf{F}_{3D} = [\mathbf{F}_{xy}, F_z]^T$ represents the composite force vector decomposed into planar and depth components, $\tau$ denotes planar torque, $M$ is object mass, and $I$ is the moment of inertia. The planar dynamics $\mathbf{F}_{xy}$ are computed using Pymunk's rigid-body engine, while $F_z$ is regulated to maintain smooth depth motion.

\paragraph{Perspective-Consistent Scaling} A key advantage of our approach over 2D methods is realistic perspective handling. As objects move along the depth axis, their apparent size changes according to perspective projection:

\begin{equation}
S(d) = S_0 \cdot \frac{f}{f + d}
\end{equation}
    
where $S_0$ is the original size at reference depth, $d$ is the depth displacement, and $f$ is the effective focal length estimated from the input image using camera calibration heuristics~\cite{fu2025geowizard}. This linear scaling approximates the perspective foreshortening effect, enabling visually coherent animations where objects naturally appear larger when approaching the camera and smaller when receding.

\paragraph{Language-guided Initialization} Our framework executes three key tasks: (1) language-guided object insertion, which introduces objects at specified locations with physically accurate initialization; (2) language-guided object removal, which eliminates target objects while maintaining physical consistency; (3) language-guided object control, which applies user-specified forces and torques in the augmented space.

\paragraph{Collision Handling} Our system employs a layer-wise collision processing pipeline. Objects are assigned to depth layers based on their current depth $d_i(t)$. Collision detection and response are computed within each layer using Pymunk's 2D engine. This design choice trades off cross-layer interaction accuracy for computational efficiency and implementation simplicity. Objects moving across layer boundaries are reassigned dynamically. While this approximation cannot handle collisions between objects at significantly different depths, it effectively captures interactions among objects in similar depth ranges, which constitute the majority of realistic scenarios. The depth-dependent scaling ensures that collision responses appear visually consistent with the objects' apparent sizes.

\subsection{Physics-guided Video Synthesis}

Our video synthesis pipeline integrates physics-based motion, depth-aware appearance modeling, and neural rendering to produce temporally coherent and physically plausible animations. The synthesis process consists of three key stages: motion-guided composition, depth-aware relighting, and physics-constrained refinement.

\noindent\textbf{Motion-guided Composition:} We create an initial video sequence by alpha-blending the warped foreground objects onto the inpainted background according to simulated transformations, with scale adjustments based on relative depth:
\begin{equation}
\hat{X}(t) = composite(B, {warp(X^i, T_i(t)) \cdot S_i(d(t))}_i)
\end{equation}
where $X^i$ represents the segmented image of the $i$-th object appearance, $B$ denotes the clean background, $T_i(t)$ represents the affine transformation matrix, and $S_i(d(t))$ is a depth-dependent scaling factor.

\noindent\textbf{Depth-aware Relighting:} To model appearance changes due to varying depths, we employ a relighting module that computes shading based on transformed albedo $\hat{A}(t)$, surface normals $\hat{N}(t)$, and estimated directional light $L$:
\begin{equation}
\tilde{X}(t) = f(\hat{X}(t), \hat{A}(t), \hat{N}(t), L, d(t))
\end{equation}

\noindent\textbf{Physics-constrained Refinement:} The relit sequence is refined through a latent diffusion model~\cite{rombach2022high} that ensures physical consistency while improving visual fidelity. We apply an adaptive fusion strategy in latent space:
\begin{equation}
z_{t-1} = (1-m)z_{t-1} + m[w(t)z_{t-1} + (1-w(t))\tilde{z}_{t-1}]
\end{equation}
where $m$ is the foreground mask and $w(t)$ regulates fusion strength.

\begin{figure*}[t]
\begin{center}
\includegraphics[width=0.99\textwidth]{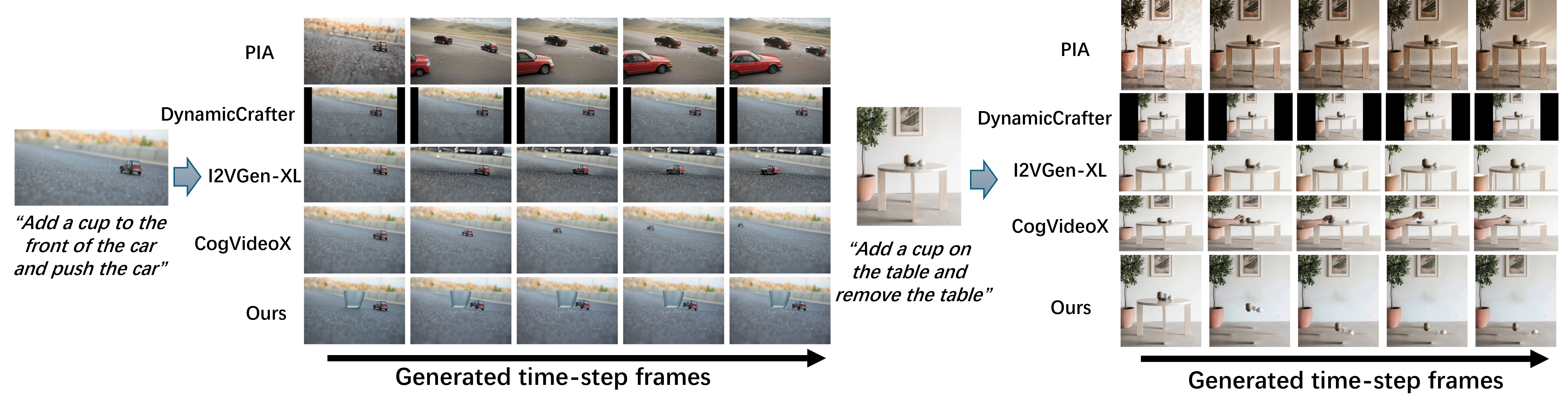}
\end{center}
\caption{Qualitative comparison of different methods on two challenging scenarios. Our method generates more realistic and coherent results with improved physics plausibility and temporal consistency compared to baseline approaches.}
\label{fig:qualitative}
\end{figure*}

\section{Experiments}

\subsection{Implementation Details}

\noindent\textbf{Scene Understanding:} Our implementation leverages several state-of-the-art models. We employ Gemini-pro-vision~\cite{team2023gemini} for language-vision understanding and physical property inference. The VLM prompt includes: (1) task description (insert/remove/control), (2) object identification requirements, (3) physical property queries (mass range: 0.1-10kg, friction: 0.1-1.0, elasticity: 0.1-0.9). We use Grounded-SAM~\cite{ren2024grounded} for instance detection and segmentation, Geowizard~\cite{fu2025geowizard} for depth and normal estimation, Clipaway~\cite{ekin2024clipaway} for background inpainting and object removal, and Diffree~\cite{zhao2024diffree} to generate objects and masks from text descriptions.

\noindent\textbf{Animation Pipeline:} We implement physics simulation using Pymunk~\cite{blomqvistpymunk} with our depth-aware extensions. The depth dimension is discretized into $L=4$ to 8 layers depending on scene complexity. We run 160 simulation steps at 30Hz and sample 16 frames uniformly to create 512$\times$512 resolution videos. The perspective scaling factor $S(d)$ is computed using Eq. 2 with focal length $f$ estimated from the depth map. For video refinement, we employ SEINE with DDIM sampling (25 steps, noise strength $s=0.5$).

\subsection{Experimental Setup}

\noindent\textbf{Dataset:} To evaluate our approach across diverse scenarios, we curate indoor and in-the-wild images with variations in illumination, viewpoints, and scene complexity. Our evaluation dataset comprises three language-guided tasks (object insertion, removal, and control), each containing 20 unique initial images. For each image, we generated 5 videos with different physical parameters and initial conditions, yielding a total of 60 source images and 300 videos.

\noindent\textbf{Baselines:} We compare against five state-of-the-art image-to-video generation methods: DynamicCrafter~\cite{xing2025dynamicrafter}, I2VGen-XL~\cite{zhang2023i2vgen}, PIA~\cite{zhang2024pia}, CogVideoX~\cite{yang2024cogvideox}, and PhysGen~\cite{liu2025physgen} as a physics-aware baseline.

\subsection{Quantitative Results}

Following prior works, we evaluate using four metrics: CLIP-Similarity~\cite{radford2021learning} to measure alignment between instructions and video frames; FID~\cite{heusel2017gans} to assess visual quality; Motion-FID to evaluate motion coherence; Motion-S to assess smoothness of generated motions.

Table~\ref{tab:quantitative} shows our method achieves superior performance across all metrics. The improved CLIP-Sim score (16.64 vs $\leq$ 16.28) indicates better instruction-video alignment, while the lower FID score (102.32 vs $\leq$ 112.83) indicates enhanced visual quality. The improvement in Motion-FID (389.64 vs $\leq$ 401.61) demonstrates that our depth-aware physical simulation generates more coherent and realistic object motions.

\begin{table}[htbp]
\caption{Quantitative comparison with baseline methods.}
\begin{center}
\begin{tabular}{lcccc}
\toprule
\textbf{Methods} & \textbf{Clip-Sim} & \textbf{FID} & \textbf{M-FID} & \textbf{M-S} \\
\midrule
PIA           & 16.17 & 215.06 & 418.51 & 0.981 \\
DynamiCrafter & 16.16 & 185.86 & 429.22 & 0.954 \\
CogVideoX     & 16.18 & 120.91 & 406.32 & 0.978 \\
I2VGen-XL     & 16.28 & 187.35 & 442.91 & 0.956 \\
PhysGen       & -     & 112.83 & 401.61 & 0.982 \\
\textbf{Ours} & \textbf{16.64} & \textbf{102.32} & \textbf{389.64} & \textbf{0.986} \\
\bottomrule
\end{tabular}
\label{tab:quantitative}
\end{center}
\end{table}

\subsection{Ablation Study}

To validate the contribution of each component, we conduct ablation experiments by removing key elements from our framework. We evaluate on 30 videos across all three tasks using the same metrics as the main comparison.

\begin{table}[h]
\caption{Ablation study results. Each row removes one key component.}
\begin{center}
\begin{tabular}{lcccc}
\toprule
\textbf{Variant} & \textbf{Clip-Sim} & \textbf{FID} & \textbf{M-FID} & \textbf{M-S} \\
\midrule
Full Method & \textbf{16.64} & \textbf{102.32} & \textbf{389.64} & \textbf{0.986} \\
w/o Depth scaling ($S(d)=1$) & 16.52 & 118.45 & 405.28 & 0.979 \\
w/o Depth motion ($v_z=0$) & 16.48 & 115.67 & 398.42 & 0.981 \\
w/o Relighting & 16.61 & 108.94 & 392.11 & 0.985 \\
w/o Language guidance & 15.93 & 105.28 & 391.87 & 0.984 \\
\bottomrule
\end{tabular}
\label{tab:ablation}
\end{center}
\end{table}

\noindent Depth scaling is the most critical component: removing it ($S(d){=}1$) causes the largest FID degradation (102.32$\rightarrow$118.45). Depth motion removal ($v_z{=}0$) similarly hurts visual quality (FID$\rightarrow$115.67). Relighting removal moderately affects FID ($\rightarrow$108.94), while removing language guidance most impacts text-video alignment (CLIP-Sim: 16.64$\rightarrow$15.93). All components contribute meaningfully to the full model.

\subsection{User Study}

We conducted a user study with 10 participants who evaluated 30 randomly selected videos based on three criteria—text-video alignment, visual quality, and physical plausibility—using a 5-point Likert scale.

As shown in Table~\ref{tab:human_eval}, our method outperforms all baselines significantly. It achieves the highest score for text alignment (4.23 vs $\leq$ 3.12), visual quality (3.83 vs $\leq$ 3.56), and most notably, physical plausibility (4.10 vs $\leq$ 3.31).

\begin{table}[htbp]
\caption{Human evaluation results (1-5 scale).}
\begin{center}
\begin{tabular}{lccc}
\toprule
\textbf{Methods} & \textbf{Text-Align} & \textbf{Visual} & \textbf{Physics} \\
\midrule
PIA           & 1.86 & 1.68 & 1.55 \\
DynamiCrafter & 1.97 & 2.02 & 1.82 \\
I2VGen-XL     & 2.21 & 2.54 & 2.16 \\
CogVideoX     & 3.12 & 3.35 & 2.63 \\
PhysGen       & -    & 3.56 & 3.31 \\
\textbf{Ours} & \textbf{4.23} & \textbf{3.83} & \textbf{4.10} \\
\bottomrule
\end{tabular}
\label{tab:human_eval}
\end{center}
\end{table}

\section{Limitations and Future Work}

While PhysLayer advances depth-aware image animation, several limitations remain. Our method is restricted to rigid body dynamics and cannot handle deformable objects, articulated structures, or complex physical phenomena such as fluid dynamics. The depth-layered collision detection prioritizes computational efficiency over complete accuracy, preventing realistic interactions between objects at significantly different depths. Our approach assumes a fixed camera viewpoint and cannot support significant view changes without full 3D scene reconstruction. From a practical perspective, the pipeline requires approximately 100--105 seconds per 16-frame video on an A6000 GPU after model initialization (with the diffusion renderer accounting for roughly 58\% of the total time), limiting real-time applications, and the reliance on VLMs for physical property estimation can introduce inaccuracies for uncommon materials. Future work will focus on extending to non-rigid motion, improving cross-layer interaction modeling, incorporating learned physical property predictors, accelerating inference for interactive applications, and supporting camera motion with multi-view consistency.

\section{Conclusion}

We introduce PhysLayer, a framework for language-guided, depth-aware layered animation of static images. Our method decomposes scenes into depth-based layers using vision models, extends 2D rigid-body physics with depth motion and perspective-consistent scaling, and integrates simulated trajectories with scene-aware relighting to produce temporally coherent and physically plausible videos. This approach provides a practical balance between physical realism and computational efficiency without requiring full 3D reconstruction. Extensive experiments demonstrate that PhysLayer outperforms existing methods across multiple metrics, achieving substantial improvements in FID (9.3\%), physical plausibility (24\%), text-video alignment (35\%), CLIP-Similarity (2.2\%), and Motion-FID (3\%), while offering flexible control over diverse animation effects through natural language guidance.

\bibliographystyle{IEEEbib}
\bibliography{icme2026references}

\end{document}